\pdfoutput=1
\documentclass[11pt]{article}

\usepackage{acl}

\usepackage[utf8]{inputenc}
\usepackage{mdframed}
\usepackage{enumitem}
\usepackage{amsmath}
\usepackage{times}
\usepackage{latexsym}
 \usepackage{graphicx}
\usepackage[T1]{fontenc}
\usepackage{longtable}
\usepackage{float}
\usepackage[utf8]{inputenc}
\usepackage{multirow}
\usepackage{booktabs}

\usepackage{microtype}

\usepackage{inconsolata}

\title{Improving LLM-based Autonomous Web Agents with Filtering}

\author{
  Zhitong Guo\textsuperscript{*} \qquad
  Jing Yu Koh\textsuperscript{*} \qquad
  Ruiyu Li\textsuperscript{*}\\[6pt]
  Carnegie Mellon University, Pittsburgh, PA, United States \\
  {\footnotesize\textsuperscript{*}{All authors contributed equally to this work and are co-first authors.}}
}
\begin{document}
\maketitle 
\thispagestyle{plain}
\pagestyle{plain}
\begingroup
\renewcommand\thefootnote{}
\footnotetext{\footnotesize{The initial version of this research was completed in 2023.}}
\addtocounter{footnote}{-1}
\endgroup
\begin{abstract}
Autonomous web agents, powered by Large Language Models (LLMs), has garnered significant attention for automating various web-based tasks with multi-step reasoning and decision-making capabilities. An open research question in the development of these agents lies in the format of the webpage input. Raw HTML source code, with its extensive and often irrelevant details, poses difficulties for LLMs with limited context windows. To address the challenge, we first reproduce baseline models such as GPT-3.5 and LLaMA-2-70B on the WebArena~\cite{webarena} benchmark, identifying common failure modes. We then proposed two retrieval strategies to filter out irrelevant context for LLM agents. We developed DeBERTa-based and T5-based models that can rank HTML elements by their relevance to the task. We finetuned them on Mind2Web trajectory data and transferred them to WebArena. Experiments show that our DeBERTa-based model enhances the success rate of the LLaMA-2-70B LLM agent on WebArena from 1.97\% to 2.96\%.  Moreover, we developed zero-shot ColBERT-based retriever, which is able to retrieve the ground-truth element with a recall of 0.52 on Mind2Web and 0.47 on WebArena.

\end{abstract}

\section{Introduction}

There has been significant recent interest in developing autonomous language model agents for the web~\cite{pmlr-v70-shi17a,LiuGPSL18,webshopyao,deng2023mind2web,gur2023realworld,webarena}. These are agents powered by Large Language Model (LLM) backbones, capable of multi-step reasoning, decision-making, and taking actions within an environment (e.g., webpages on the Internet). Successfully building such agents would allow us to automate a wide variety of web-based tasks, improving human efficiency and augmenting capabilities.

\begin{figure*}[t]
\vskip 0.2in
\begin{center}
\centerline{\includegraphics[width=\textwidth]{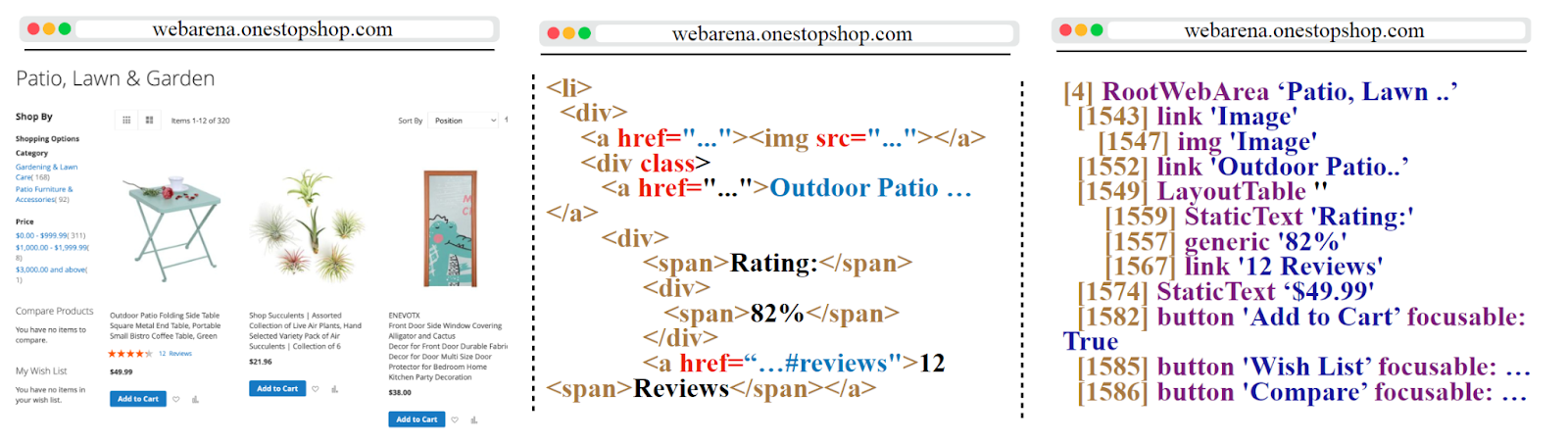}}
\caption{Observation Space for WebArena: screenshot (left), HTML DOM tree (middle), and accessibility tree (right). Figure reproduced from \citet{webarena}.}
\label{fig:observations}
\end{center}
\vskip -0.2in
\end{figure*}

In order to develop a capable autonomous web agent, the format of the input observation for the LLM is a crucial first consideration. Currently, LLM baselines in WebArena (e.g., GPT-3.5 and GPT-4 reasoning baselines) ingest the information processed from the HTML source of the current screen of the webpage as the ``observation'' of the current state of the webpage (Figure~\ref{fig:observations}). However, the information provided in HTML source code is often extremely detailed and verbose, and significantly fills up the LLM's context window (often with information that is irrelevant to the task at hand). Information from the HTML source (e.g. style attributes, JavaScript code snippets, CSS-related code blocks) is often unrelated to the objective of a language model agent and serve as distractors, hindering the performance of the agent. In addition, due to the dynamic nature of most websites nowadays, a lot of the HTML elements are not actually visible in the rendered state of the web page (e.g., a navigation drawer where the user could hide and show information inside through clicking or hovering).

Although the raw HTML source code is the most complete observation of the state of the website, it may not be the most suitable for LLMs and thus may not result in the best agent execution accuracy. In WebArena, they utilize better input data representations through extracting the simplified accessibility tree representation with the Chromium browser.

As shown in Figure~\ref{fig:observations}, the accessibility tree (right), compared with the raw HTML source (middle), is a stripped-down version with only key hierarchical information, as well as inferred semantic information of what each element’s key role and information is about. This format of input representation is heuristically created from raw HTML and can already save a lot of context length burden for LLMs. It also has clearer types of elements like button, link, textbox, or tables, etc. 

For similar motivations as the previous project idea, a long HTML source tends to be an issue for the limited context length that LLM can consume. As a result, a more focused and high-efficiency input data representation is desired. In WebArena \cite{webarena}, several techniques are already employed to shrink down the size, e.g. using an accessibility tree instead of raw HTML, only containing elements present in the current user viewport, etc. Along with this goal, we aim to investigate ways of finding relevant portions of the current input representation based on the user intent. For example, if the user intent is about finding flight tickets to New York, then ideally the input representation to the LLM should contain only the flight search part of a travel website. We hypothesize that with this in-page retrieval with intents, we could potentially have a more efficient input. 

There are also previous efforts for denoising HTML or filtering out irrelevant information. For example, Mind2Web \cite{deng2023mind2web} and WebAgent \cite{gur2023realworld} both train specific filtering or extractive summarization models for this goal of reducing input. However, without training data, we want to explore other possible ways, e.g. retrieval for selecting the most important part of the page, given user intent. 

As part of this work, we reproduced results of baseline models on the WebArena benchmark. In particular, we re-ran the GPT-3.5 baseline model and achieve comparable success rate to their paper. In addition, we also implemented inference with the LLaMA-2-70B~\cite{touvron2023llama} model, and report its performance on WebArena. Based off these results, we perform a comprehensive quantitative and qualitative analysis, highlighting common failure modes of the model when performing certain types of tasks.

In addition to reproducing these results, we also investigated several directions towards improving the WebArena LLM agents:
\begin{itemize}
    \item We compared different retrieval strategies, such as using ColBERT \cite{ColBERT} and DeBERTa \cite{he2021deberta} for retrieving the top elements for the accessibility tree representation; moreover, we explored whether a DeBERTa model fine-tuned on high-qualities Mind2Web data, with its observation space mapped into WebArena observation space, can be leveraged for agent’s better performance in a more complex and dynamic environment, WebArena.
    \item We conducted extensive experiments and demonstrated that our retrieval strategy improves the LLaMA-2-70B model when applied to WebArena tasks, increasing success rate on the test set from \textbf{1.97\% to 2.96\%}.
    \item We proposed possible methods for subquery generation and planning for the agent with GPT-4 that can be generalized to open-domain websites in addition to the simulated web environments. Appendix \ref{appendsg} shows an example of our subquery generation.
\end{itemize}

\section{Related Work}
\subsection{Web Agents Controlled by Natural Language}
Previous research~\cite{pmlr-v70-shi17a,LiuGPSL18,webshopyao,yao2023react,deng2023mind2web,gur2023realworld,webarena} has made efforts towards build autonomous agents for Web tasks controlled by natural language. However, there still remains a substantial room to improve agent performance on dynamic and diverse web environments. 

Some works~\cite{deng2023mind2web,gur2023realworld} focus on creating more informative text to input into LLMs for web tasks. Mind2Web \cite{deng2023mind2web} ranks and filters the raw HTML of real-world websites with a compact language model, such as DeBERTa, and identifies the top-k potential candidates. Subsequently, it prompts the process of element selection for the LLM as a multi-choice question-answering (QA) problem, offering five candidates for each iteration. It's worth noting that Mind2Web develops its method based on DOM tree inputs. To improve the input data representation for LLMs, We plan to build our agent taking in accessibility tree inputs, which is more concise yet informative. Moreover, Mind2Web consists of static website snapshots. In contrast, our proposed approach can be applied to dynamic real-world websites.

To address the challenge of HTML code lengths often exceeding the context limits of most LLMs, WebAgent~\cite{gur2023realworld} 
decomposes language instructions into canonical sub-instructions, summarizes long HTML
documents into task-relevant snippets using a LLM, and provide the sub-instruction and HTML snippets as input to a second LLM to generate programs for performing web tasks.  WebAgent also introduced the HTML-T5 model. HTML-T5 is enriched with global and local attention mechanisms, and is specifically fine-tuned to make HTML code more concise. Furthermore, WebAgent initiates its planning process with a set of predefined sub-instructions, in contrast with the WebArena agent which dynamically plans subsequent steps based on the evolving content of the web page following each interaction. Consequently, WebArena agents require real-time screenshot capture and inference capabilities to adapt to changing web environments. 

Several other works~\cite{webshopyao,LiuGPSL18} develop reinforcement learning or imitation learning agents for web tasks. WebShop~\cite{webshopyao} leverages recent pre-trained language models for representing and generating text to learn search and choose on a realistic and interactive web environment. However, WebShop primarily focuses on the item-selection task, and does not handle a diverse range of tasks as WebArena~\cite{webarena} does. Moreover, the tasks within WebShop do not involve any updates or modifications to the website itself, making the tasks relatively straightforward in nature. \citet{LiuGPSL18} also adopts a much simpler representation of real-world environmental complexities than \citet{webarena} does. In contrast, our work go beyond the breadth and depth of their tasks to tackle more complex and varied task execution on WebArena~\cite{webarena}.

For these reasons, we develop our work based on the WebArena~\cite{webarena} benchmark. On the agent side, it develops LLM-prompt-based web agents where both the input observations and the predicted actions are all in text. The agents simply untilze LLM's reasoning abilities and prompt engineering techniques. We will provide more details and our reproduction in later sections.

On the dataset side, WebArena is a fully functional web environment spanning four distinct domains: e-commerce, social forum discussions, collaborative software development, and content management. It emulates a diverse array of real-world web tasks and offers a collection of natural language commands and their associated responses. WebArena encompasses 241 intent templates and 812 intents, categorically classified into site-navigation tasks, information-seeking tasks, and content configuration tasks. WebArena is purposefully structured to facilitate sequential inference for web agents. In this framework, at each step, the agent is presented with an observation space, tasked with selecting an action from the available action space. Subsequently, the action leads to the emergence of a new state and a fresh observation. The observation space encompasses the web URL and its associated content, which comprises (1) a screenshot capturing the current viewport, (2) the HTML DOM tree, and (3) the HTML accessibility tree. Concurrently, the action space is delineated as a set of possible actions, encompassing interactions such as clicking, hovering, typing, and more. Additionally, the agent is tasked with identifying a web element for interaction, which can be specified either by its Element ID within the Document Object Model (DOM) tree or by its on-screen coordinates.

%\subsection{Natural Language Command and Control}
%Other works~\cite{yao2023react} \cite{}.

\subsection{Pre-trained Language Model and Fine-tuning}
Recent advances in pre-trained large language models build off the Transformer~\cite{vaswani2017attention} architecture. Models such as BERT~\cite{Devlin2019BERTPO}, have boosted the performance on many natural language processing tasks, including information retrieval (IR) and knowledge reasoning. Fine-tuning further empower LLMs to solve a wider range of tasks by adapting them to a particular task or dataset, gaining domain-specific understanding.

DeBERTa~\cite{he2021deberta} is a Transformer-based LM that improves the BERT and RoBERTa models using two novel techniques. First, each word in DeBERTa is represented using two vectors that encode its content and position, respectively, and a disentangled attention mechanism is used to allow the model to capture contents and relative positions of word pairs. Second, an enhanced mask decoder is used to incorporate absolute positions in the decoding layer to predict the masked tokens for model pre-training. Moreover, the paper achieves better training stability of fine-tuning DeBERTa to a downstream NLP task by introducing a virtual adversarial training algorithm.

ColBERT~\cite{ColBERT} is a ranking model that adapts LMs for efficient retrieval. ColBERT fine-tunes two different pre-trained BERT models to separately encode Query and Document text into contextual embeddings. Based on the embeddings, ColBERT then computes a relevance score for each query-document pair through late interaction and performs ranking. By delaying the query–document interaction, It preserves the effectiveness of BERT and reduces the cost of IR.

Recently, successful scaling of LLMs has substantially improved model performance and exhibited emergent abilities in addressing complex NLP tasks. Among existing LLMs, GPT-3.5~\cite{NEURIPS2020_1457c0d6} and LLaMA-2-70B~\cite{touvron2023llama} are decoder-only Transformer models. The models have vast number of parameters, making fine-tuning for specific tasks very challenging. An alternative to finetuning is to craft few-shot in-context examples and provide them as input context during inference. \citet{NEURIPS2020_1457c0d6} showed that LLMs are capable of learning from these examples to improve over zero-shot performance. Several existing LLM agents (such as the baseline results we reproduce from WebArena~\cite{webarena}) are prompt-based, and include several examples to enable agents to improve when provided with accessibility tree inputs.

\section{Reproduced Baseline Results}
We first reproduced GPT-3.5 results from the original WebArena paper~\cite{webarena} to verify our implementation. In addition, we also ran LLaMA-2-70B~\cite{touvron2023llama} on the WebArena benchmark (which was not reported in the original paper). We found these results valuable as it allows us to measure the performance of open sourced models, which despite being less expressive and capable compared to GPT-3.5, provides more options in finetuning and modeling extensions. Our results are reported in Table~\ref{tab:results}, and the trajectories can be viewed on Zeno ML platform linked in our repository.

\subsection{GPT-3.5}
We run the GPT-3.5-turbo baseline through the original WebArena source code by following the official WebArena instructions, with the command provided at \footnote{\url{https://github.com/ZhitongGuo/11711-webarena}}. The temperature is set to 1.0, the top-$p$ parameter is set to 0.9, the maximum state transition is set to 30, which are aligned with the settings in the original paper. The prompting methods we used combined CoT and without UA hint (Unachievable hint) mentioned in the WebArena paper, which explicitly instructs the agent to stop if it believes the task is impossible to perform.

\subsection{LLaMA-2-70B}
We run the LLaMA-2-70B~\cite{touvron2023llama} model by accessing its model weights through its official HuggingFace page\footnote{\url{https://huggingface.co/meta-llama/Llama-2-70b-hf}}. In order to run the LLaMA-2-70B model, we served it using the \texttt{lti-llm-deployment}\footnote{\url{https://github.com/neulab/lti-llm-deployment/tree/main}} library, which allows us to host a HuggingFace model on a server to enable API calls. Then, we ran the original WebArena evaluation code, replacing the API calls to OpenAI with API calls to \texttt{lti-llm-deployment}. All parameters were kept the same as the GPT-3.5 baseline, except that we set decoding temperature to 0.6 and the \texttt{top\_p} to 0.95, as we found that the LLaMA model tends to produce bad generation results when temperature is set to 1.0 (as it is set in the GPT-3.5 WebArena baselines). The script to launch the LLaMA-2-70B API endpoint is provided in our GitHub repo.

\begin{table*}[t]
\centering
\begin{tabular}{lccc}
\toprule
\textbf{Model}          & \textbf{Site} & \textbf{Success Rate (SR) ($\uparrow$)} & \textbf{Overall SR ($\uparrow$)} \\ 
\midrule % Replaces the second \hline
\multirow{5}{*}{\texttt{gpt-3.5}} & gitlab        & 3.33\%                     & \multirow{5}{*}{6.65\%}     \\
                         & reddit        & 3.77\%                     &                              \\
                         & shopping\_admin & 4.94\%                  &                              \\
                         & map           & 9.17\%                     &                              \\
                         & shopping      & 13.37\%                    &                              \\ 
\midrule
\multirow{5}{*}{\texttt{gpt-3.5 + DeBERTa + k = 50\%}} & gitlab        & 6.21\%                     & \multirow{5}{*}{5.84\%}     \\
                         & reddit        & 2.83\%                     &                              \\
                         & shopping\_admin & 2.20\%                  &                              \\
                         & map           & 9.17\%                     &                              \\
                         & shopping      & 7.48\%                    &                              \\ 
\midrule
\multirow{5}{*}{\texttt{LLaMA-2-70B}}   & gitlab        & 3.06\%                    & \multirow{5}{*}{1.97\%}      \\
                         & reddit        & 0.92\%                     &                              \\
                         & shopping\_admin & 0.00\%                  &                              \\
                         & map           & 1.80\%                     &                              \\
                         & shopping      & 3.66\%                     &                              \\ 
\midrule
\multirow{5}{*}{\texttt{LLaMA-2-70B + DeBERTa + k = 50}}   & gitlab        & 1.53\%                    & \multirow{5}{*}{2.34\%}      \\
                         & reddit        & 0.0\%                     &                              \\
                         & shopping\_admin & 2.20\%                  &                              \\
                         & map           & 1.79\%                     &                              \\
                         & shopping      & 5.21\%                     &                              \\ 
\midrule
\multirow{5}{*}{\texttt{LLaMA-2-70B + DeBERTa + k = 50\%}}   & gitlab        & 2.56\%                    & \multirow{5}{*}{2.96\%}      \\
                         & reddit        & 0.88\%                     &                              \\
                         & shopping\_admin & 1.10\%                  &                              \\
                         & map           & 6.25\%                     &                              \\
                         & shopping      & 4.69\%                     &                              \\ 
\bottomrule % Replaces the last \hline
\end{tabular}
\caption{Comparison of GPT 3.5 and LLaMA model success rates on WebArena sites.}
\label{tab:results}
\end{table*}

\section{Methods}
\subsection{Finetuning DeBERTa and T5 as retrievers} \label{sec:deberta}

\begin{figure}[t]
    \centering
    \includegraphics[width=1.0\linewidth]{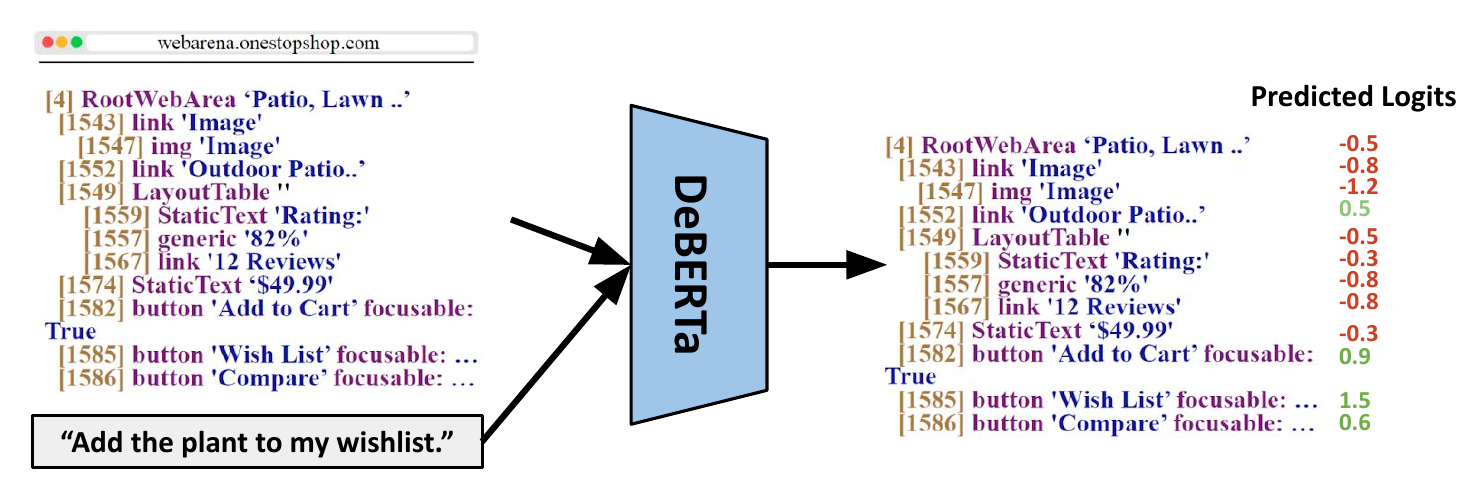}
    \caption{We fine-tune a DeBERTa model on Mind2Web~\cite{deng2023mind2web} to classify whether each HTML element is relevant to the task or not. During inference, we rank the logits of the elements in descending order and keep the top-$k$ elements.}
    \label{fig:deberta_inference}
\end{figure}

As the WebArena data does not contain groundtruth labels, we fine-tuned them on the Mind2Web~\cite{deng2023mind2web} training data. Mind2Web contains HTML data and corresponding groundtruth element IDs that represent the next element in the human trajectory for accomplishing the task. We preprocessed the Mind2Web HTML elements similarly to the WebArena accessibility tree inputs. For each element, we labeled it as 1 if it is the groundtruth element, and 0 otherwise. Due to the significant class imbalance (only 1 element per page is labeled as 1), we oversampled the positive class to balance it as 50-50 0 and 1 labels during training time. We train the model by minimizing the binary cross entropy loss.

\subsection{Zero-shot ColBERT retriever}
The ColBERT retrieval system is comprised of an indexer and a searcher. In the indexing phase, each document $d$ within a webpage viewport is transformed into a corresponding set of vectors. Concurrently, the query $q$ is encoded into a multi-vector representation. The similarity between a query $q$ and a document $d$ is computed by aggregating the maximal cosine similarities across the query tokens, specifically through the application of "MaxSim" operations. This procedure involves identifying the highest cosine similarity between each query token embedding and the embeddings of all tokens in the passage. The similarity score $S_{q, d}$ is formally defined as:
$$
S_{q, d}=\sum_{i=1}^N \max _{j=1}^M Q_i \cdot D_j^T
$$

Then, we construct the observation with two heuristics based on the retrieved results. \\
\textbf{Adjacency window.} For each top 5 element in the retrieved results, select the neighboring 20 elements for each (5 before and 5 after) and concatenate the resulting window.\\
\textbf{Bounded window.} Select the topmost and the bottom-most elements in retrieved results, and construct the observation as upper and lower bounds.

\subsection{Augmenting LLMs with Filtering}
%\todo{@Zhitong can you add the WebArena plots you made here?}

During evaluation, we first run the DeBERTa retrieval model on all the elements in the accessibility tree of the current observation (Figure~\ref{fig:deberta_inference}). For each HTML element on the page, we pass it to the fine-tuned DeBERTa model, and extract its logits. Then, we sort the logits of all elements on the page by descending order, and only retain the top 50\% highest scored elements. These elements are used as the input to the downstream LLM agent, thus reducing context length by half on average.

\subsection{Collecting Data and Choosing Parameters}
To add step-wise trajectory data, we cleaned 207 WebArena trajectories by ourselves and further compared the performance of DeBERTa and zero-shot Colbert on it. In our cleaned trajectories, some of the trees are not bounded by viewports and thus much longer than those in WebArena that are bounded. 

\begin{figure}
    \centering
    \includegraphics[width=0.95\linewidth]{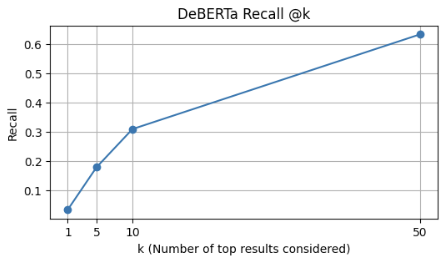}
    \caption{DeBERTa on cleaned WebArena trajectories}
    \label{arenakDeBERTa}
\end{figure}

\begin{figure}
    \centering
    \includegraphics[width=0.95\linewidth]{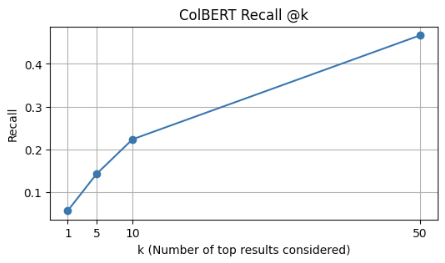}
    \caption{ColBERT on cleaned WebArena trajectories}
    \label{arenakColBERT}
\end{figure}

Figure \ref{arenakDeBERTa} and \ref{arenakColBERT} shows that when the number of top results we retained increases, both DeBERTa and ColBERT achieve higher recall of task success. DeBERTa outperforms ColBERT consistently. In Table \ref{tab:results}, we also experimented with $k = 50$ and $k = 50\%$ and $k = 50\%$ gives better results. So we set $k = 50\%$ for our following experiments. In the future, we want to develop better ways of choosing how many top web elements to retain.

\section{Results and Discussions} \label{sec:results}
% In this section, we analyze several common failure modes of the stronger GPT-3.5 baseline, and suggest possible directions to remediate this in future work (i.e., Assignment 4).

\subsection{Retrieval Results and Model Comparison}
\begin{figure}[ht]
    \centering
    \includegraphics[width=0.95\linewidth]{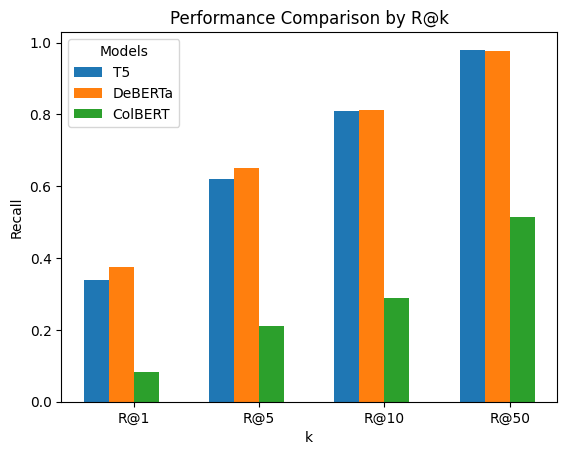}
    \caption{Validation results of the three models we tried on Mind2Web.}
    \label{retrivalmodel}
\end{figure}

We measured the performance of our fine-tuned T5, DeBERTa, and 0-shot ColBERT models on the validation set of Mind2Web. As shown in Figure \ref{retrivalmodel}, we find that the fine-tuned models achieve good Recall@k overall. In particular, 81\% of the time, the groundtruth element is in the top 10 results. For comparison, DeBERTa and T5 outperform ColBERT greatly. Since DeBERTa also outperforms ColBERT slightly, we transferred DeBERTa to WebArena. 

\subsection{Quantitative Analysis}
\paragraph{Baseline results} We first discuss the baseline results reproduced from the regular WebArena benchmark. The breakdown of the site-level success rate is shown in Table~\ref{tab:results}. The overall success rate for gpt-3.5 with CoT and without UA hint, 6.65\% is slightly better (but comparable) with the 6.41\% success rate as listed in WebArena \cite{webarena}, and we attribute the minor difference due to the non-determinism of the gpt-3.5 model.

We also observe that the success rates are significantly higher for the map and shopping environments for gpt-3.5, with success rates of 9.17\% and 13.37\% respectively. This suggests that the tasks in these environments may be simpler for the model, as it requires less complex reasoning compared to more complicated environments such as the shopping\_admin site, which involves interacting with the Adobe Magento admin portal\footnote{\url{https://experienceleague.adobe.com/docs/commerce-admin/start/admin/tools/admin-dashboard.html}}, which is significantly more complex than regular websites, involving interactive charts, dropdowns, and tables. It is also likely out of distribution with respect to the regular websites that gpt-3.5 was trained on. For LLaMA-2-70B, we observed similar trends, with the shopping site performing significantly higher than the other WebArena websites. Interestingly, both LLaMA-2-70B and gpt-3.5 achieved a success rate of 3.33\% on gitlab.

\paragraph{Results with DeBERTa filtering model}

We ran both the GPT-3.5 and LLaMA-2-70B models, using the trained DeBERTa model (Sec.~\ref{sec:deberta}) to filter out irrelevant parts of the accessibility tree representation before it is passed to the LLM. This is done similarly to training, where we run the DeBERTa model over every HTML element on the page, and keep the top-$k$ with the highest logits. This results in a significantly reduced accessibility tree representation, allowing us to retain just the most relevant elements (as scored by the trained DeBERTa model).

The results are summarized in Table~\ref{tab:results}. We observed that filtering the inputs with DeBERTa improves the overall results, from 1.97\% to 2.96\%. 

By error analysis, we recognize two possible reasons for why retriever doesn't improve the performance of gpt-3.5. Since gpt-3.5 has much longer context length than LLaMA, gpt-3.5 is able to take in most of the complete accessibility tree input. After our retriever filters the input, the tree layout is disrupted. So our model lose the layout information that the original model obtains. Besides, the recall of DeBERTa on WebArena is just 0.63, implying that some groundtruth elements are filtered out by our retriever.

\paragraph{Site-level breakdown}

\begin{figure}[h]
    \centering
    \includegraphics[width=5cm]{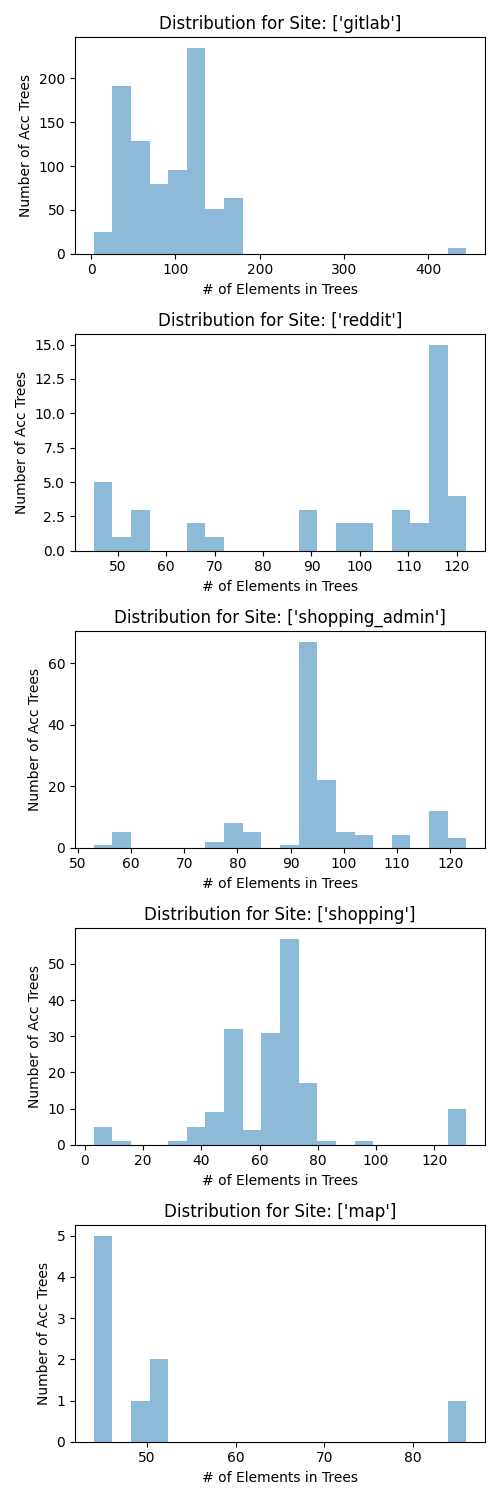}
    \caption{Site Level Breakdown of distributions}
    \label{WebArena Analysis}
\end{figure}

Analyzing the distributions of the number of elements in the accessibility trees (Figure \ref{WebArena Analysis}), a correlation emerges between the quantity of these elements and the relative complexity of tasks across different websites. Specifically, tasks on the shopping and map sites are characterized by a lower count of elements, averaging around 50 and 70 per observation, respectively. This lower complexity correlates with higher initial success rates for both GPT-3.5 and LLaMA-2-70B models on these sites. In contrast, tasks on GitLab, Reddit, and a shopping administration interface present a higher level of difficulty, predominantly featuring around 100 elements in their accessibility trees. Notably, GitLab tasks occasionally exhibit a significant spike in number of HTML elements, with elements exceeding 400 in certain instances. Our retriever improved the performance of gpt-3.5 agent on GitLab tasks and the performance of LLaMA agent on shopping admin and shopping tasks. 

\paragraph{Deteriorated performance on certain tasks} We observed quantitatively and qualitatively that performance deteriorates on several tasks in the benchmark as well. We attribute this to false-negatives from the filtering stage, where an essential element is removed by the filtering model. This leads to error compounding which the downstream LM is unable to correct, as the relevant elements are removed from its input. Training the DeBERTa model with a different training objective than cross-entropy loss in order to maximize recall may be a way to alleviate this problem. Intuitively, we may be more accepting of false-positives for our proposed pipeline, as the downstream LM would have a chance to process these elements and correct it. False-negatives are significantly more damaging, and training the classification model to minimize false-negatives would be a promising direction for future exploration.

\subsection{Qualitative Analysis} \label{sec:qualitative_analysis}

\begin{figure*}[!ht]
    \centering
    \begin{minipage}{0.35\textwidth}
        \includegraphics[width=\linewidth]{img_50pc_llama/task36_screenshot.png} 
    \end{minipage}
    \begin{minipage}{0.29\textwidth}
        \includegraphics[width=\linewidth]{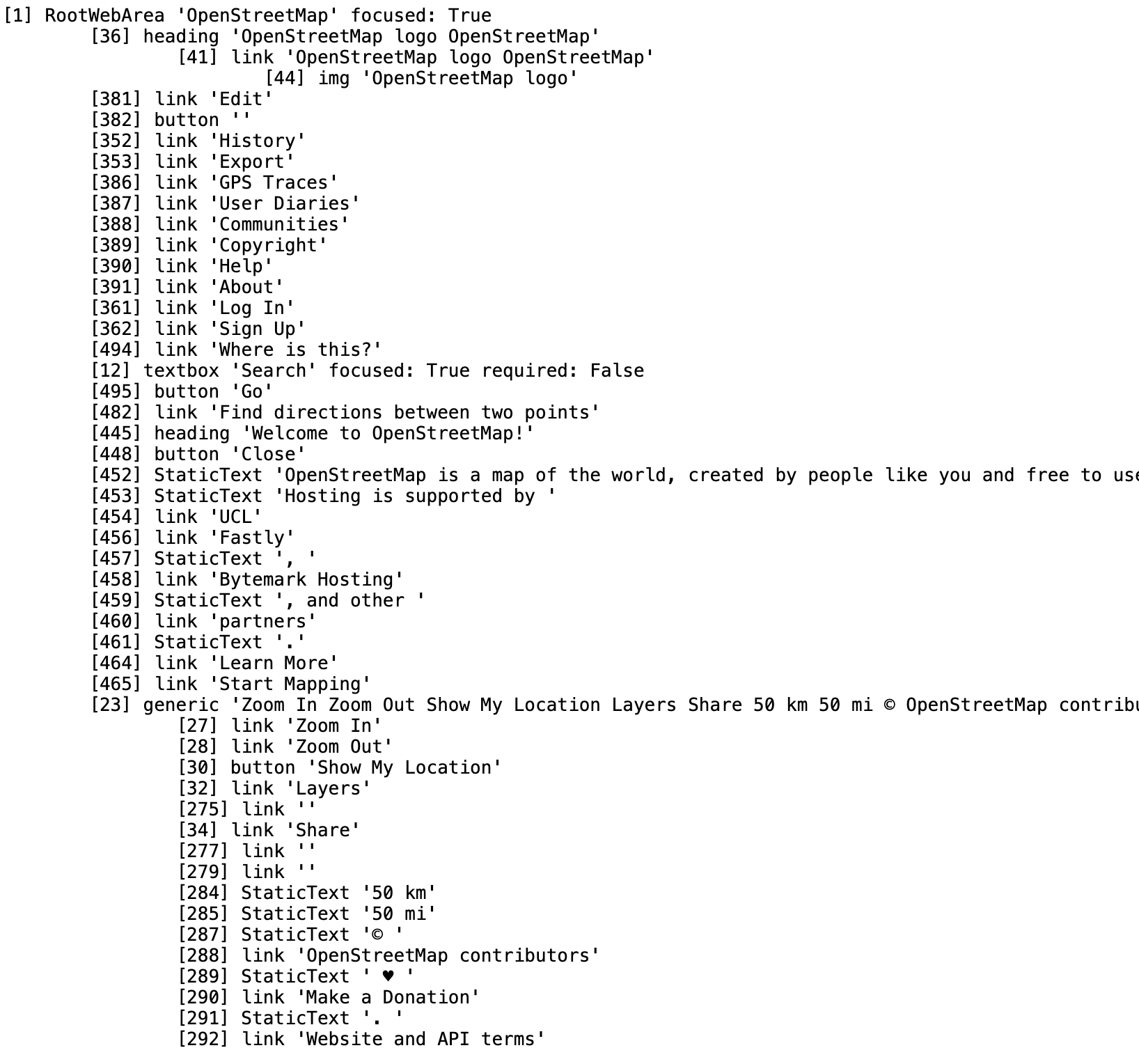}
    \end{minipage}
    \begin{minipage}{0.29\textwidth}
        \includegraphics[width=\linewidth]{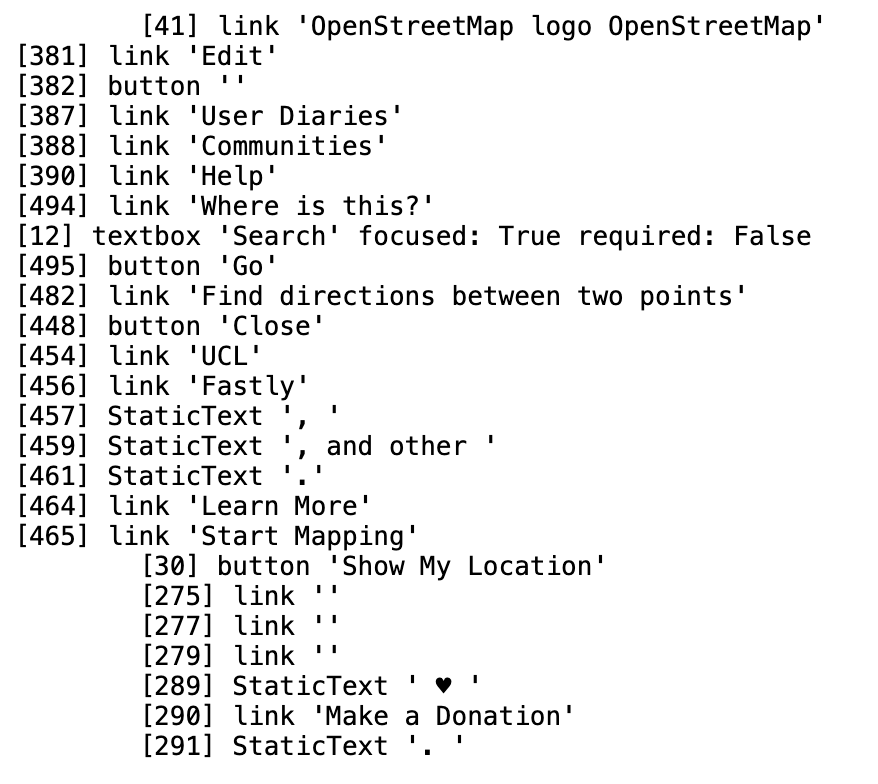}
    \end{minipage}
    \caption{Task 36 screenshot (left), raw accessibility tree with 1594 characters (center), and filtered accessibility tree to keep 50\% of elements, resulting in 658 characters (right). Filtering the accessibility tree to keep just the top-50\% items allows the model to focus on the relevant parts of the context. The original LLaMA-2-70B fails on this task, while the model with the filtered tree succeeds.} \label{fig:task36_50pc}
\end{figure*}

\begin{figure*}[!ht]
    \centering
    \begin{minipage}{0.35\textwidth}
        \includegraphics[width=\linewidth]{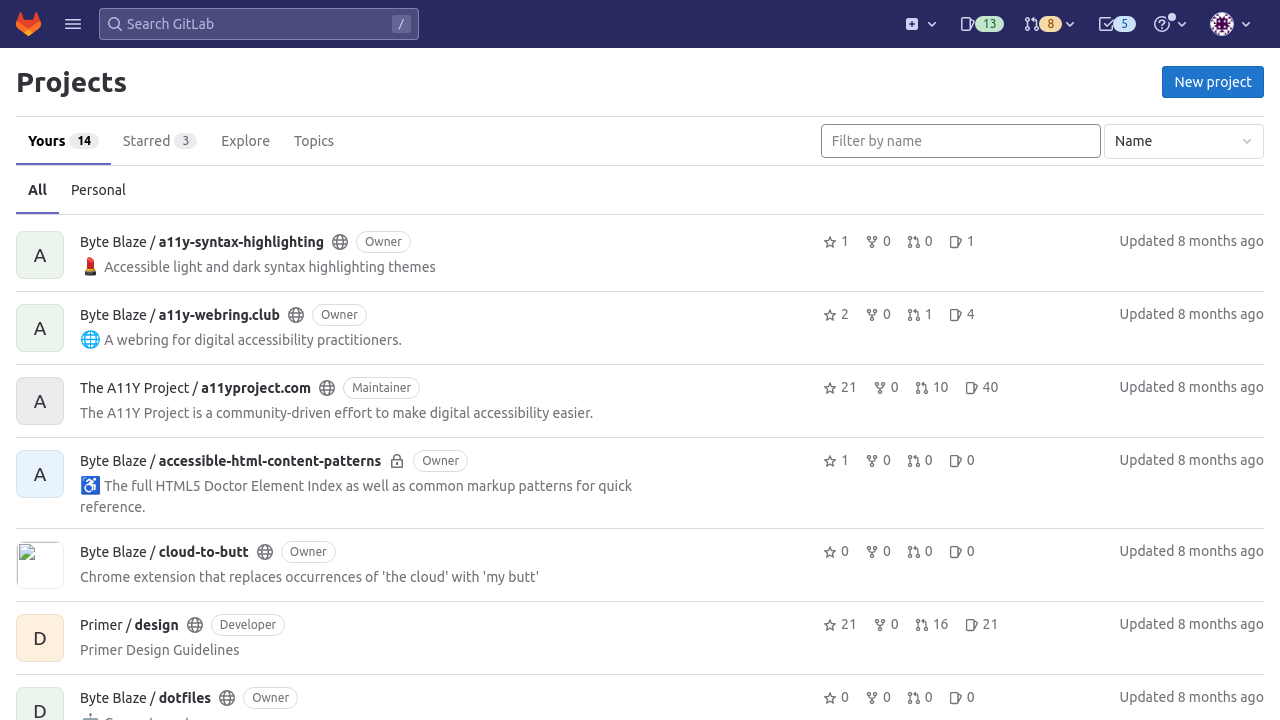} 
    \end{minipage}
    \begin{minipage}{0.29\textwidth}
        \includegraphics[width=\linewidth]{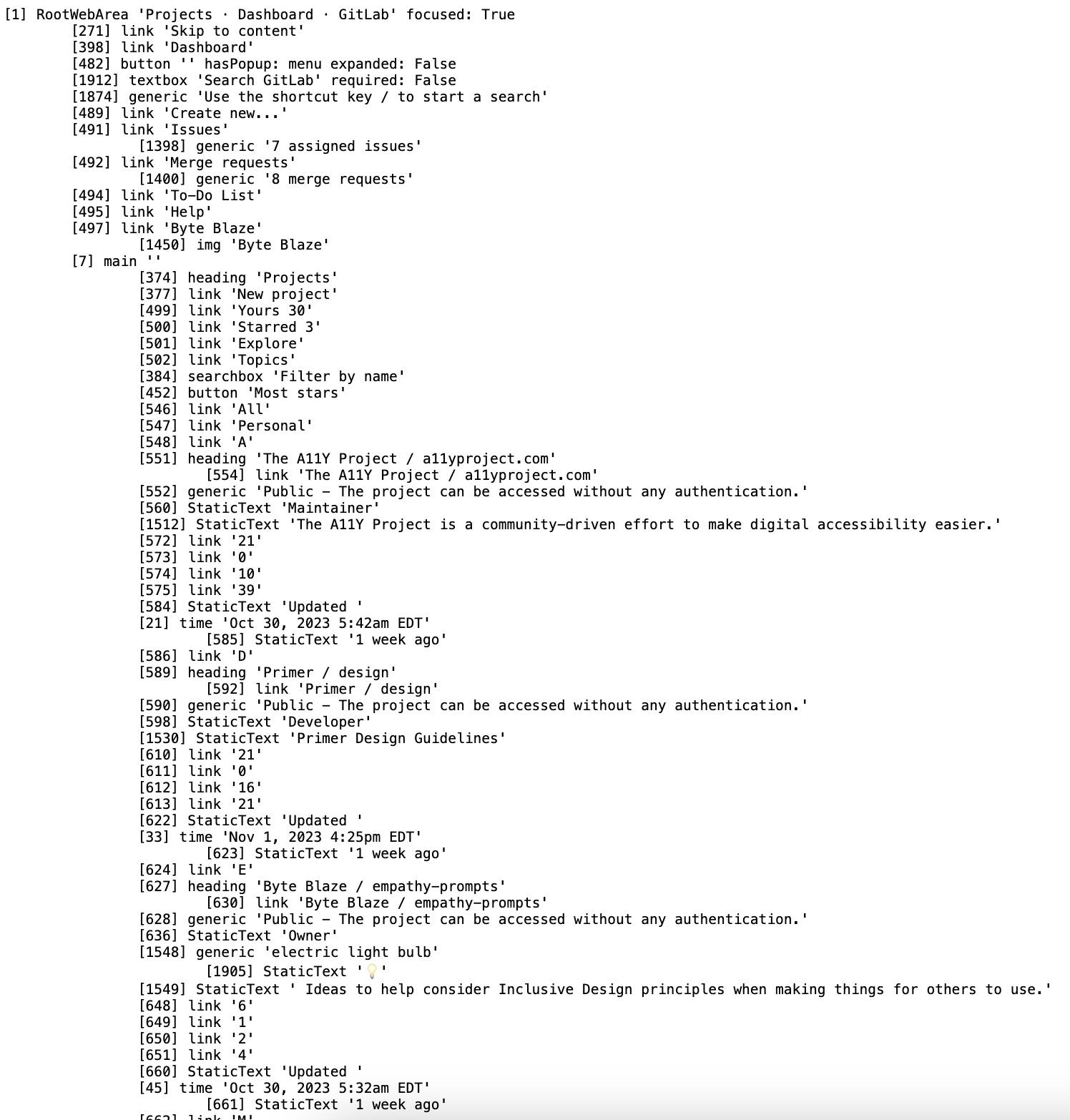}
    \end{minipage}
    \begin{minipage}{0.29\textwidth}
        \includegraphics[width=\linewidth]{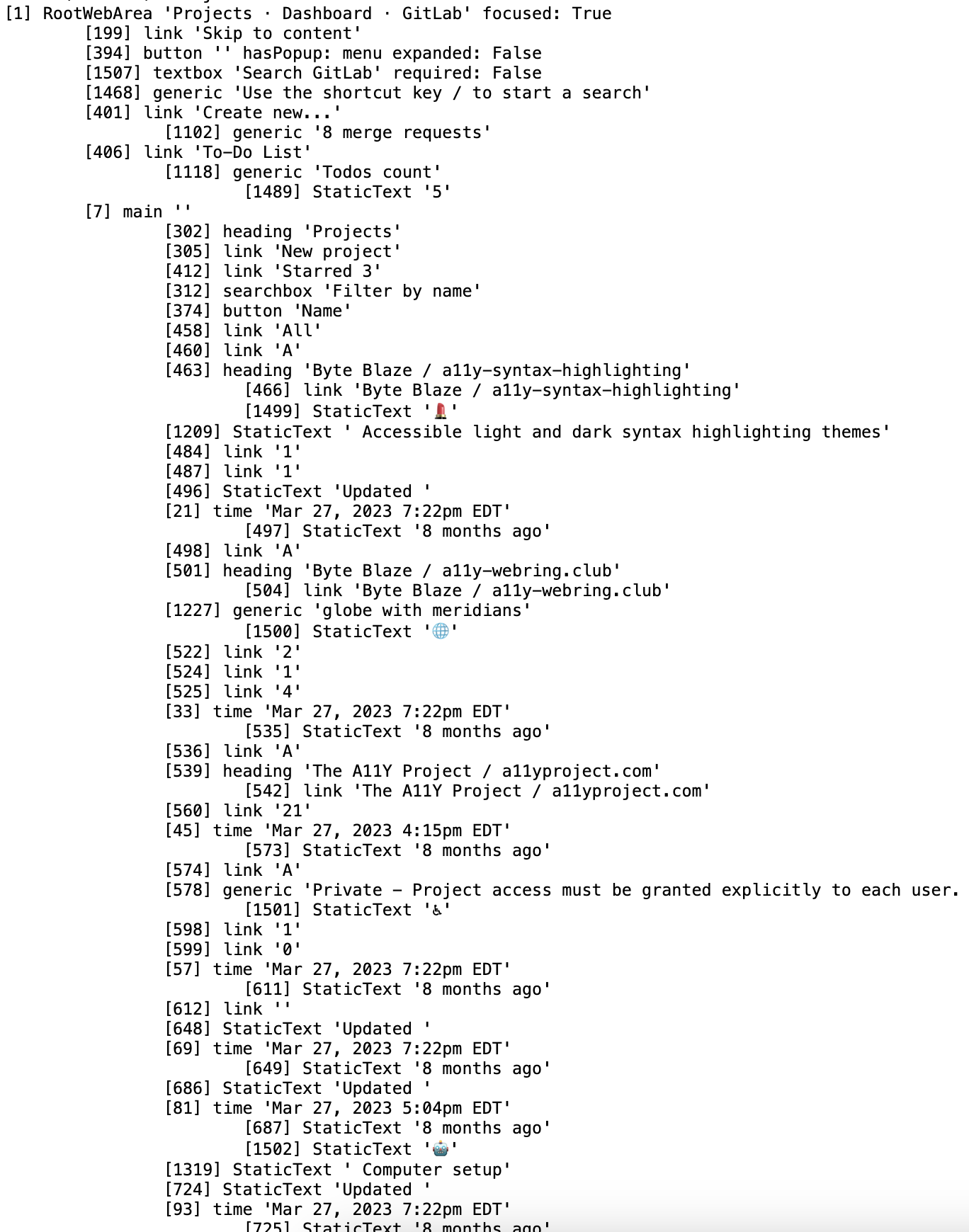}
    \end{minipage}
    \caption{Task 132 screenshot (left), raw accessibility tree with 4319 characters (center), and filtered accessibility tree to keep 50\% of elements, resulting in 2138 characters (right). The original LLaMA-2-70B fails on this task, while the model with the filtered tree succeeds.} \label{fig:task132_50pc}
\end{figure*}

\begin{figure*}[!ht]
    \centering
    \begin{minipage}{0.35\textwidth}
        \includegraphics[width=\linewidth]{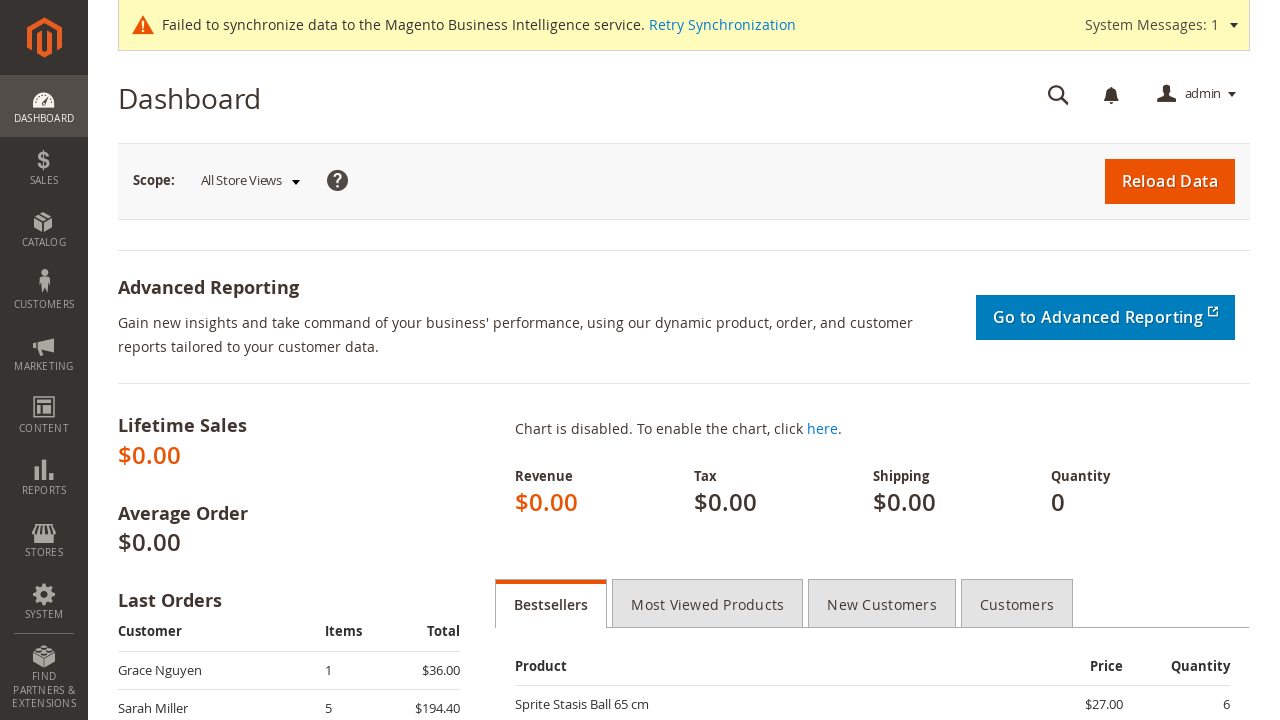} 
    \end{minipage}
    \begin{minipage}{0.29\textwidth}
        \includegraphics[width=\linewidth]{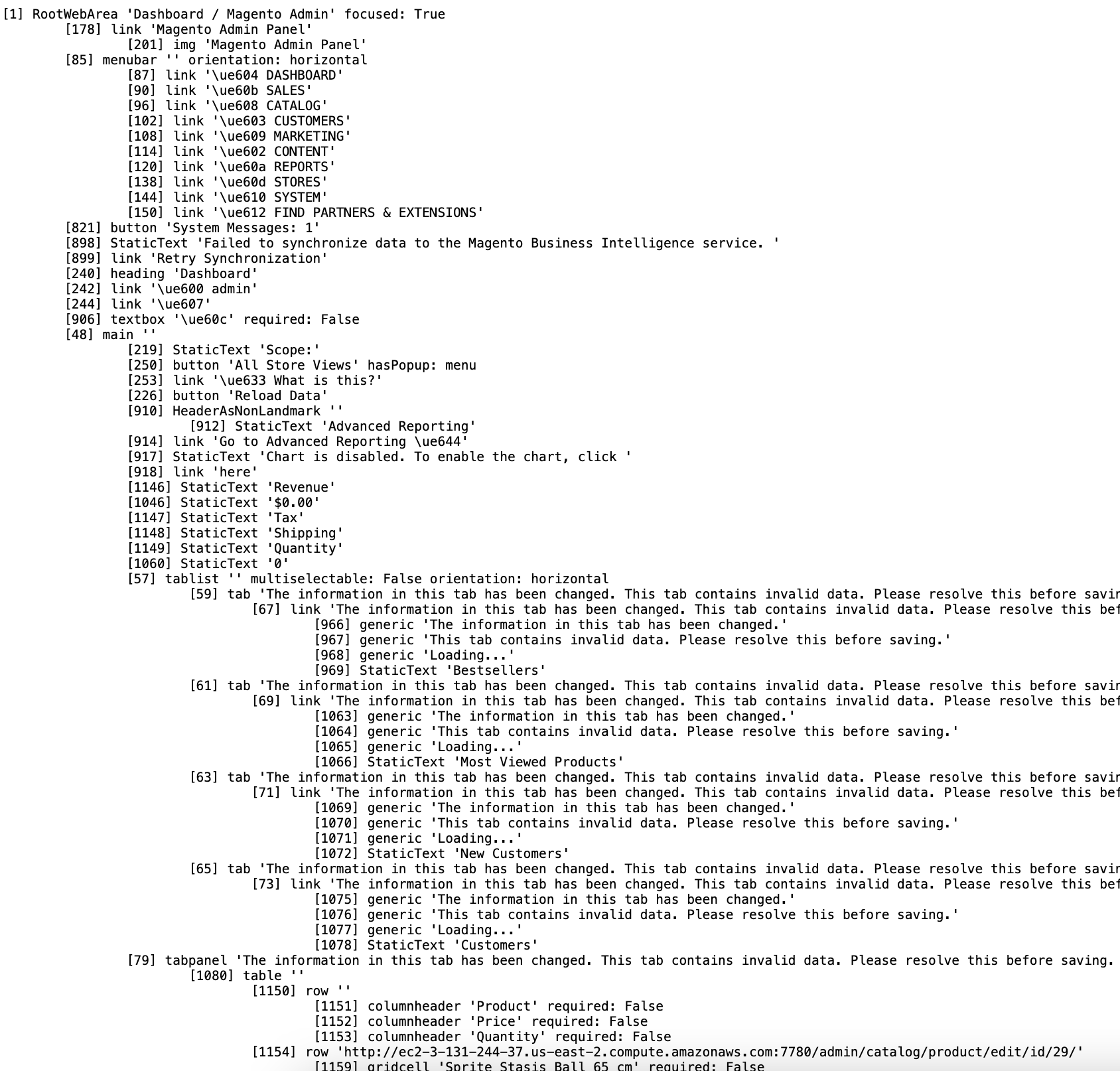}
    \end{minipage}
    \begin{minipage}{0.29\textwidth}
        \includegraphics[width=\linewidth]{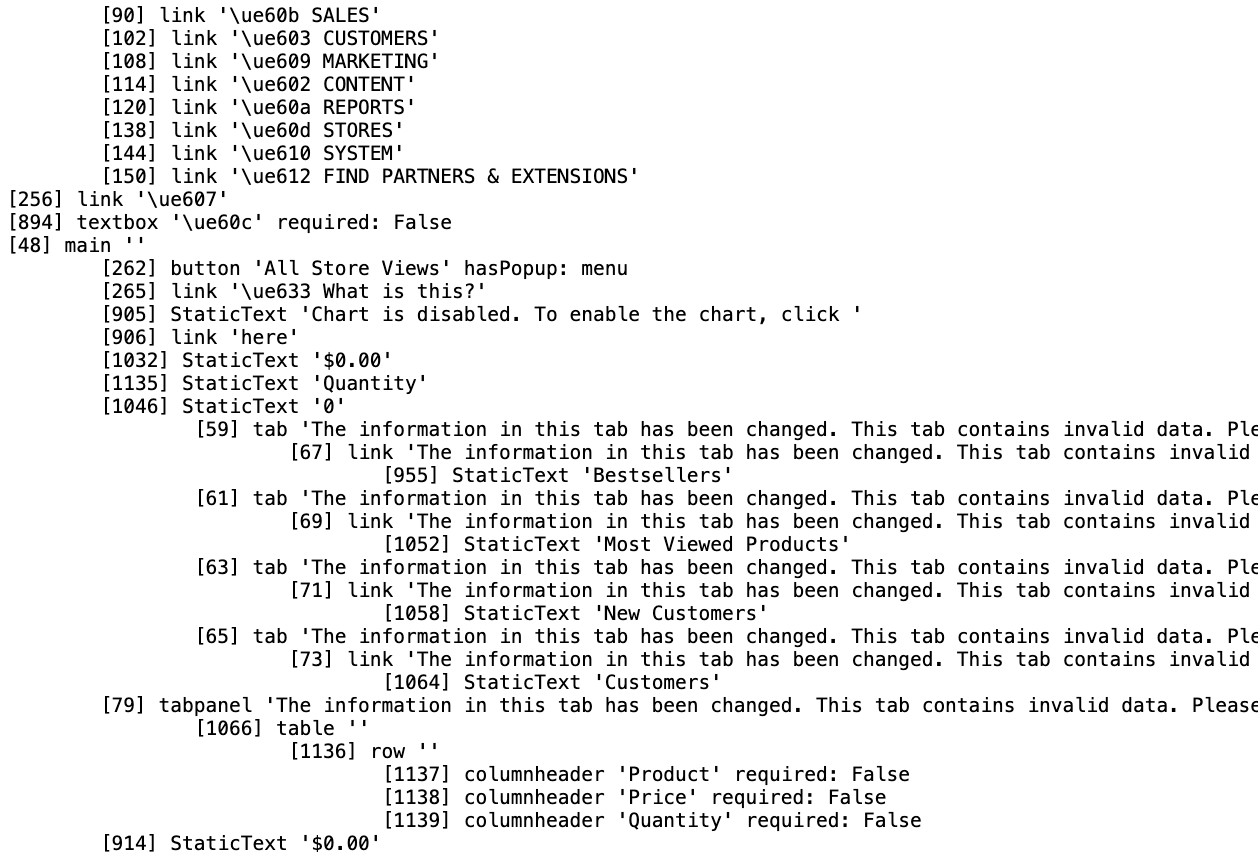}
    \end{minipage}
    \caption{Task 247 screenshot (left), raw accessibility tree with 5131 characters (center), and filtered accessibility tree to keep 50\% of elements, resulting in 2841 characters (right). The original LLaMA-2-70B fails on this task, while the model with the filtered tree succeeds.} \label{fig:task247_50pc}
\end{figure*}

\paragraph{LLaMA-2-70B Agent} We show several examples of the original accessibility tree, and the filtered accessibility tree in Figures~\ref{fig:task36_50pc}, \ref{fig:task132_50pc}, and \ref{fig:task247_50pc}. We observe that in several examples, the DeBERTa model is capable of preserving most of the useful elements in the page, allowing the downstream LM to in principle operate without loss of useful information. These results also suggest that it appears to benefit LMs with shorter context length, as they are often unable to fit the full accessibility tree within their context window (for LLaMA, this is 4K tokens, while for GPT-3.5, this is 16K tokens).

\paragraph{GPT-3.5 Agent} Figure \ref{fig:task152} provides an example of how our retriever \textbf{resolves the repetitive action problem} the and enhance the agent's performance. The example considers Task 152 as a case in point \ref{success-traj}. The retriever effectively excluded a majority of elements deemed irrelevant, including certain StaticText components and links like ``sharing'' and ``Make a Donation''. Prior to the application of filtering, the action recorded was `type [599] [Schenley park]', executed thrice consecutively, resulting in premature termination of the process. Post-filtering, the action executed by the agent is accurate, leading to a sequence of actions that makes sense, and finally the correct answer.

\section{Conclusion and Future Work}
In conclusion, we proposed two retrieval strategies to filter out irrelevant context for downstream LLM agents. We developed DeBERTa and T5 retrievers, fine-tuned them on Mind2Web trajectory data, and transferred them to WebArena. We also tested ColBERT zero-shot on Mind2Web and WebArena HTML data. On Mind2Web, the fine-tuned models achieve good
Recall@k overall. \textbf{81\%} of the time our models are able to catch the groundtruth element in their top 10 results. On WebArena, we experimented with both close-source gpt-3.5 and open-source LLaMA-2. Experiments show that our proposed DeBERTa-based model \textbf{improved} overall success rate of a LLaMA-2-70B agent on the test set from 1.97\% to 2.96\%. 

For future, we plan to fine-tune our retriever-enhanced agents on the trajectory data we just cleaned to improve their web navigation capabilities. We will also try subquery generation and planning via prompt engineering. This will allow the LLM-based agent to generalize to open-domain websites in addition to the simulated web environments. We show an example of our subquery generation in Appendix \ref{appendsg}.

\newpage % The empty page
\bibliography{custom}
\newpage
\appendix
\onecolumn
\section{Appendix} \label{sec:appendix}

\subsection{Improvement of Repetitive Actions}
\label{success-traj}
\begin{figure*}[ht]
    \centering

    \begin{minipage}{0.49\textwidth}
        \includegraphics[width=\linewidth]{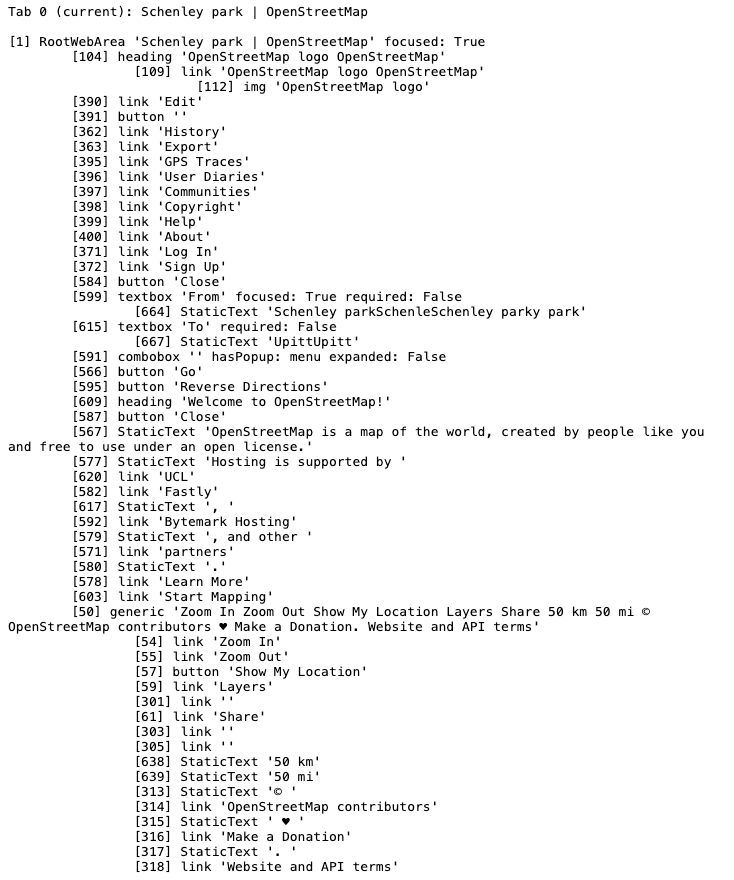}
    \end{minipage}
    \begin{minipage}{0.49\textwidth}
        \includegraphics[width=\linewidth]{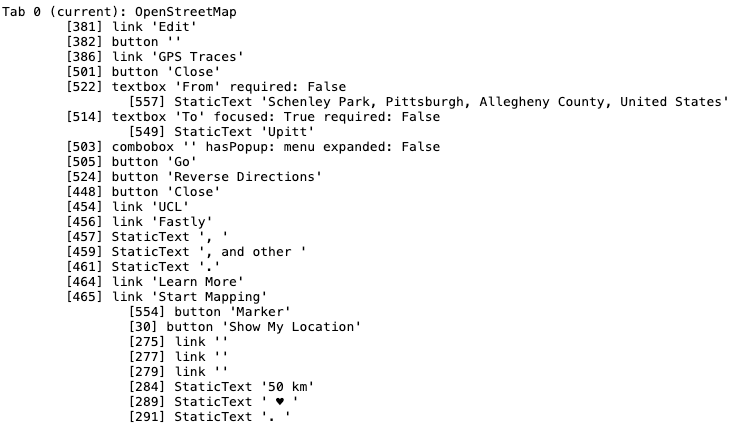} 
    \end{minipage}

    \begin{minipage}{0.49\textwidth}
        \includegraphics[width=\linewidth]{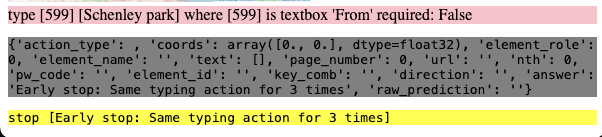}
    \end{minipage}
    \begin{minipage}{0.49\textwidth}
        \includegraphics[width=\linewidth]{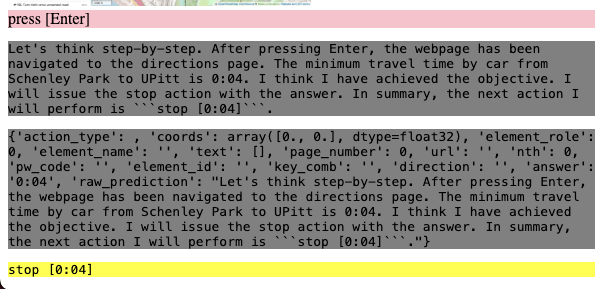}
    \end{minipage}

    \caption{Task 152, the observation before (left) and after filtering (right). The action predicted by GPT-3.5 before was \texttt{type [599] [Schenley park]}, which was repeated thrice, leading to early stopping. This is improved after filtering of the context window, enabling the model to arrive at the correct answer.} 
    \label{fig:task152}
\end{figure*}

\subsection{Subquery Generation}
\label{appendsg}
\begin{mdframed}
\textbf{Prompt:}
Task: Compare the difference in time for walking and driving route from Randyland to Carnegie Mellon University
Observation: \\
\text{[1]} RootWebArea 'OpenStreetMap' focused: True\\
	\text{[400]} link 'Edit'\\
	\text{[342]} link 'History'\\
	\text{[343]} link 'Export'\\
	\text{[409]} link 'Help'\\
	\text{[410]} link 'About'\\
	\text{[351]} link 'Log In'\\
	\text{[352]} link 'Sign Up'\\
	\text{[469]} button 'Close'\\
	\text{[473]} textbox 'From' required: False\\
	\text{[479]} textbox 'To' required: False\\
	\text{[484]} combobox '' hasPopup: menu expanded: False\\
		\text{[486]} menuitem 'Bicycle (OSRM)' selected: True\\
		\text{[487]} menuitem 'Car (OSRM)' selected: False\\
		\text{[488]} menuitem 'Foot (OSRM)' selected: False\\
	\text{[489]} button 'Go'\\
	\text{[491]} button 'Reverse Directions'\\
		\text{[261]} button 'Show My Location'\\
		\text{[262]} link 'Layers'\\
		\text{[273]} StaticText '500 m'\\
		\text{[274]} StaticText '2000 ft'\\
Let's think step by step. List bullet points of sub tasks for the task show above.\\

\textbf{Agent:}
To compare the difference in time for a walking and driving route from Randyland to Carnegie Mellon University using OpenStreetMap, follow these sub-tasks:\\
1.  Input Starting Point : Enter "Randyland" into the 'From' textbox.\\
2.  Input Destination : Enter "Carnegie Mellon University" into the 'To' textbox.\\
3.  Select Travel Mode for First Route :\\
    - Choose the "Foot (OSRM)" option from the combobox to set the mode of travel to walking.\\
4.  Search for Walking Route :\\
    - Click the 'Go' button to generate the walking route.\\
5.  Note Walking Route Time :\\
    - Once the route is displayed, note the estimated time for the walking route.\\
6.  Reset Travel Mode for Second Route :\\
    - Re-enter "Randyland" and "Carnegie Mellon University" in the respective textboxes.\\
    - Choose the "Car (OSRM)" option from the combobox to set the mode of travel to driving.\\
7.  Search for Driving Route :\\
    - Click the 'Go' button again to generate the driving route.\\
8.  Note Driving Route Time :\\
    - Once the route is displayed, note the estimated time for the driving route.\\
9.  Compare Times :\\
    - Compare the noted times for walking and driving to understand the difference in duration between the two modes of transportation.\\
10.  Optional Steps :\\
    - If interested, also take note of the route details like distance, route directions, and any significant landmarks or turns for a more comprehensive comparison.
\end{mdframed}

\end{document}